\documentclass[runningheads]{llncs}

\newcommand {\ent} {\mathrel{{\scriptstyle\mid\!\sim}}}

\newcommand{\tip}{{\bf T}}

\newcommand{\be}{\begin{enumerate}}
\newcommand{\ee}{\end{enumerate}}

\newcommand{\hide}[1]{}

\def \cases{\left \{\begin{array}{l}}
\def \endcases{\end{array}\right .}

\newcommand {\ri} {\rightarrow}

\newcommand {\bes} {\begin{description}}
\newcommand{\ens} {\end{description}}
\newcommand {\la} {\langle}
\newcommand {\ra} {\rangle}

\newcommand {\beq} {\begin{quote}}
\newcommand {\enq} {\end{quote}}
\newcommand {\bit} {\begin{itemize}}
\newcommand {\enit} {\end{itemize}}

\newenvironment{pozz}{\color{black}}{\color{black}}
\usepackage{color}
\usepackage{times}
\usepackage{helvet}
\usepackage{courier}
\usepackage{times}
\usepackage{undertilde}
\usepackage{graphicx}
\usepackage{latexsym}
\usepackage{graphicx}
\usepackage{color}
\usepackage{amssymb}
\usepackage{colortbl}
\usepackage{amsmath}
\usepackage{microtype}
\usepackage{amsmath}
\usepackage{tikz}

\def \ri{\rightarrow}

\begin{document}
\bibliographystyle{plain}

\title{Many-valued Argumentation, Conditionals  and \\ a Probabilistic Semantics for Gradual Argumentation}



\author{Mario Alviano \inst{1}  \and Laura Giordano \inst{2}  \and  Daniele Theseider Dupr{\'{e}} \inst{2} }

\institute{
 Universit\`a della Calabria,  Italy
\and
Universit\`a del Piemonte Orientale, Italy 
}

%

\authorrunning{ }
\titlerunning{ }

 \maketitle
 

\begin{abstract}
In this paper we propose 
a general approach to define a many-valued preferential interpretation of gradual argumentation semantics. The approach allows for conditional reasoning over arguments and boolean combination of arguments, with respect to a class of gradual semantics, through the verification of graded (strict or defeasible) implications over a preferential interpretation. 
As a proof of concept, in the finitely-valued case, an Answer set Programming approach  is proposed for conditional reasoning  in a many-valued argumentation semantics of weighted argumentation graphs.
The paper also develops and discusses a probabilistic semantics for gradual argumentation, which builds on the many-valued conditional semantics.

\end{abstract}


\section{Introduction}

Argumentation is one of the major fields in non-monotonic reasoning (NMR) which has been shown to be very relevant for decision making and for explanation \cite{Toni2019}.
The relationships between preferential semantics of commonsense reasoning \cite{Pearl90,whatdoes,GeffnerAIJ1992,BenferhatIJCAI93}
and argumentation semantics 
 are very strong  \cite{Dung95,GeffnerAIJ1992,Weydert2013,Greco2020,IsbernerFLAIRS2020}.
 
%
In particular, the relationships between some multi-preferential semantics for weighted conditional knowledge bases with typicality and gradual argumentation semantics  
\cite{CayrolJAIR2005,Janssen2008,Dunne2011,Leite2013,Amgoud2017,BaroniRagoToni2018,Amgoud2019} 
have been recently investigated \cite{WorkshopAI3_short,NMR2022}.
This has given rise to some new gradual argumentation semantics stemming from fuzzy preferential semantics for conditionals. 


This paper develops a general approach to define a preferential interpretation of an argumentation graph under a gradual semantics, to allow for {\em defeasible reasoning over the argumentation graph}, by formalizing conditional properties of the graph in a many-valued logic with typicality, i.e., a many-valued propositional logic in which arguments play the role of propositional variables, and a typicality operator is allowed, inspired by the typicality operator proposed  in the Propositional Typicality Logic  \cite{BoothCasiniAIJ19}. 
The operator allows for the definition of {\em defeasible implications} of the form $\tip(A_1) \rightarrow  A_2$, meaning that ``normally argument $A_1$ implies argument $A_2$", in the sense that
``in the typical situations where $A_1$ holds, $A_2$ also holds". 
The truth degree of such implications can be determined with respect to a preferential interpretation defined from a set of labellings of an argumentation graph, according to a chosen (gradual) argumentation semantics. They correspond to conditional implications $\alpha \ent \beta$ in the KLM approach \cite{KrausLehmannMagidor:90,whatdoes}. More precisely, the paper considers {\em graded defeasible inclusions} of the form $\tip(\alpha) \rightarrow  \beta \geq l$, meaning that  ``normally argument $\alpha$ implies argument $\beta$ 
with a degree at least $l$", where $\alpha$ and $\beta$ can be {\em boolean combination of arguments}.
%
The satisfiability of such inclusions in a {\em multi-preferential} interpretation of an argumentation graph (wrt. a given  semantics), exploits the preference relations over labellings, relations which are associated to arguments.


As a proof of concept, we consider a {\em finitely-valued} variant of  the gradual semantics  presented in  \cite{NMR2022}, called the $\varphi$-coherent semantics.
For the case when the truth degree set is $\mathcal{ C}_n =\{0, \frac{1}{n},\ldots, \frac{n-1}{n}, 1\}$, for some integer $n \geq 1$, 
an Answer set Programming approach  for conditional reasoning over an argumentation graph is developed for the $\varphi$-coherent semantics, based on the idea of encoding  the labellings of an argumentation  graph as Answer Sets.  The defeasible implications
satisfied in a preferential interpretation are determined by reasoning about preferred answer sets, namely, the answer sets maximizing the admissibility value of some arguments. 

The definition of a preferential interpretation $I^S$ associated to a gradual semantics $S$, sets the ground for the definition of a probabilistic interpretation of gradual semantics $S$. In Section \ref{sec:probabilistic_semantics}, for the gradual semantics with domain of argument valuation in the unit real interval $[0,1]$, we propose a probabilistic argumentation semantics, which builds  on a gradual semantics and is inspired by Zadeh's probability of fuzzy events.

\section{Many-valued coherent, faithfull and $\varphi$-coherent semantics for weighted argumentation graphs} \label{sec:varphi_labellings}

In this section we 
generalize the gradual argumentation semantics proposed in \cite{WorkshopAI3_short,NMR2022}, 
the $\varphi$-coherent semantics, whose definition is inspired to some recently studied semantics of conditional knowledge bases \cite{JELIA2021,ICLP22}.
As a proof of concept, the $\varphi$-coherent semantics will be used in Section \ref{sec:ASP} for conditional reasoning over an argumentation graph. 

We let the {\em domain of argument valuation} $\mathcal{S}$  to be either the real unit interval $[0,1]$ or the finite set $\mathcal{ C}_n =\{0, \frac{1}{n},\ldots, \frac{n-1}{n}, 1\}$, for some integer $n \geq 1$.
This allows to develop the notions of {\em many-valued coherent, faithfull and $\varphi$-coherent labellings} for weighted argumentation graphs, which include both the infinitely and the finitely-valued case. 

Following \cite{WorkshopAI3_short},
we let a {\em weighted argumentation graph} to be a triple 
$G=\la \mathcal{ A}, \mathcal{ R}, \pi \ra$, where $\mathcal{ A}$ is a set of arguments, $\mathcal{ R} \subseteq \mathcal{ A} \times \mathcal{ A}$  
and $\pi: \mathcal{ R} \rightarrow  \mathbb{R}$.
This definition of weighted argumentation graph is similar to {\em weighted argument system} in \cite{Dunne2011}, but here we admit both positive and negative weights, while only positive weights representing the strength of attacks are allowed in   \cite{Dunne2011}. In this notion of weighted argumentation graph, a pair $(B,A) \in \mathcal{ R}$ is regarded as a {\em support} of argument $B$ to argument $A$  when the weight $\pi(B,A)$ is positive and as an {\em attack} of argument $B$ to argument $A$  when the weight $\pi(B,A)$ is negative. This leads to bipolar argumentation, which is well-studied in argumentation literature \cite{AmgoudCayrol2004,BaroniRagoToni2018,MossakowskiN2018,PotykaAAAI21}. 
The argumentation semantics 
introduced below deals with positive and negative weights in a uniform way. 
A notion of basic strength is not introduced, even though it could be added.

Given a weighted argumentation graph $G=\la \mathcal{A}, \mathcal{ R}, \pi \ra$,  a {\em many-valued labelling of $G$} is a function $\sigma: \mathcal{ A} \rightarrow \mathcal{S}$
which assigns to each argument an {\em acceptability degree} in $ \mathcal{S}$. 

Let $\mathit{ R^{-}(A)=}$ $\mathit{\{B \mid (B,A) \in \mathcal{ R}}\}$. 
When $\mathit{R^{-}(A)= \emptyset}$, argument $A$ has neither supports nor attacks.
For $G=\la \mathcal{ A}, \mathcal{ R}, \pi \ra$ and a labelling $\sigma$, we introduce a  {\em weight $W^G_{\sigma}$ on $\mathcal{ A}$} as a partial function $W^G_\sigma : \mathcal{ A} \rightarrow  \mathbb{R} $, assigning a positive or negative support (relative to labelling $\sigma$) to all arguments $A_i \in \mathcal{ A}$ such that $\mathit{R^{-}(A_i) \neq \emptyset}$, as follows:
 \begin{align}\label{weight_of_arguments}
	W^G_\sigma(A_i)  & = \sum_{(A_j,A_i) \in \mathcal{ R}} \pi(A_j,A_i)  \; \sigma(A_j)
\end{align} 
When $\mathit{R^{-}(A_i) = \emptyset}$, $W^G_\sigma(A_i)$ is let undefined.

We exploit this notion of weight of an argument to define some different argumentation semantics for a graph $G$, by extending also to the finitely-valued case the semantics in \cite{WorkshopAI3_short}.

\begin{definition}\label{def:labellings}
Given a weighted graph $G=\la \mathcal{ A}, \mathcal{ R}, \pi \ra$ and $\sigma: \mathcal{ A} \rightarrow \mathcal{S}$ a labelling, we say that:
\begin{itemize}
\item
$\sigma$ is a {\em coherent labelling of $G$} 
if, for all arguments $A,B \in \mathcal{ A}$ s.t. $\mathit{R^{-}(A)\neq \emptyset}$ and $\mathit{ R^{-}(B)\neq \emptyset}$,
$$\sigma(A) < \sigma(B) \;  \iff \; W^G_\sigma(A) < W^G_\sigma(B); $$

\item
$\sigma$ is a {\em faithfull labelling of $G$} 
if, for all arguments $A,B \in \mathcal{ A}$ s.t. $\mathit{R^{-}(A)\neq \emptyset}$ and $\mathit{ R^{-}(B)\neq \emptyset}$,
$$\sigma(A) < \sigma(B) \;  \Rightarrow \; W^G_\sigma(A) < W^G_\sigma(B); $$

\item
given a function $\varphi: \mathbb{R} \rightarrow \mathcal{S}$,
$\sigma$ is a {\em $\varphi$-coherent labelling of $G$} 
if, for all arguments $A\in \mathcal{ A}$,
s.t. $\mathit{ R^{-}(A) \neq \emptyset}$, 
\begin{equation} \label{cond_phi_coherence}
\sigma(A) =  \varphi(W^G_\sigma(A)) 
\end{equation}  
\end{itemize}
\end{definition}
Observe that definition of $\varphi$-coherent labelling of $G$ is defined through a set of equations, as in Gabbay's equational approach to argumentation networks \cite{Gabbay2012}.
The notion of  $\varphi$-coherent labelling requires the range of function $\varphi$ to be $\mathcal{S}$.

The notions of coherent, faithfull and $\varphi$- coherent labelling of a weighted argumentation graph $G$ do not put constraints on the labelling of arguments which do not have incoming edges, provided the constraints on the labellings of all other arguments can be satisfied, depending on the semantics considered.

A many-valued $\varphi$-coherent labelling of a weigthed argumentation graph $G$ can be proven to be a coherent labelling or a faithfull labelling of $G$, under suitable conditions on  function $\varphi$, a result which extends the one in \cite{WorkshopAI3_short}. 
\begin{proposition} \label{prop:phi_coherent_labellings}
Given weighted graph $G=\la \mathcal{ A}, \mathcal{ R}, \pi \ra$:
(1) A coherent labelling $\sigma: \mathcal{ A} \rightarrow \mathcal{S}$ of $G$  is a faithfull labelling of $G$;
(2) if $\varphi$ is a {\em monotonically non-decreasing} function, a $\varphi$-coherent labelling $\sigma$ of $G$ is  a faithfull labelling of $G$;
(3) if $\varphi$ is {\em monotonically increasing},  a $\varphi$-coherent labelling $\sigma$ of $G$ is  a coherent labelling of $G$.
\end{proposition}

In \cite{WorkshopAI3_short,NMR2022} it has been shown that, for labellings with values in $[0,1]$, the notion of $\varphi$-coherent labelling relates to the framework of gradual semantics studied by Amgoud and Doder  \cite{Amgoud2019}, 
by considering a slight extension of Amgoud and Doder's {\em gradual argumentation framework} so to deal with both positive and negative weights, to capture the strength of supports and attacks.
The notion of bipolar argumentation has been widely studied in the literature, 
e.g. in \cite{AmgoudCayrol2004,BaroniRagoToni2018,MossakowskiN2018}.
Potyca \cite{PotykaAAAI21} has considered an extension of the bipolar argumentation framework  QBAFs by Baroni et al. 
which also includes the strength of attacks and supports.

As observed in \cite{WorkshopAI3_short}, 
since MultiLayer Perceptrons 
can be mapped to weighted conditional knowledge bases, they can as well be seen as a weighted argumentation graphs, with positive and negative weights, under the proposed semantics. In this view, 
$\varphi$-coherent labellings correspond to stationary states of the network, where each unit in the network is associated to an argument, synaptic connections (with their weights) correspond to attacks/supports, and the activation of units can be regarded as the values of the corresponding arguments in a labelling.
This is in agreement with previous work on the relationship between argumentation frameworks and neural networks, 
first investigated by Garcez, Gabbay and Lamb \cite{GarcezArgumentation2005} and more recently by Potyca \cite{PotykaAAAI21}. 

While we refer to \cite{WorkshopAI3_short,NMR2022} for a description of the relationships of the $\varphi$-coherent semantics with the gradual semantics studied by Amgoud and Doder  \cite{Amgoud2019},
 let us generalize the $\varphi$-coherent semantics,
by considering different functions $\varphi_i: \mathcal{ A} \rightarrow \mathcal{S}$, one for each argument $A_i$, rather than a single function $\varphi$ for all arguments. 
This generalization allows, for instance, to capture MLPs in which different activation functions are associated to different layers.
\begin{definition}\label{def:phi-labellings-S}
Given a weighted graph $G=\la \mathcal{ A}, \mathcal{ R}, \pi \ra$ and a function $\varphi_i: \mathbb{R} \rightarrow  \mathcal{S}$, for each $A_i \in \mathcal{ A}$, 
a labelling $\sigma: \mathcal{ A} \rightarrow  \mathcal{S}$ of the graph $G$ is a {\em $\varphi$-coherent labelling of $G$} 
if, for all arguments $A_i\in \mathcal{ A}$ s.t. $\mathit{ R^{-}(A_i) \neq \emptyset}$, 
$\sigma(A_i) =  \varphi_i(W^G_\sigma(A_i))$.
\end{definition}

Note that restricting the domain of argument valuation to the finite number of values in $ \mathcal{C}_n$, allows considering finitely-valued labellings which approximate infinitely-valued ones 
considered in \cite{WorkshopAI3_short,NMR2022}.
As a special case, for $\mathcal{S}=\mathcal{ C}_n$ with $n=1$ or $n=2$, one gets notions of {\em two-valued} and  {\em three-valued} semantics and, for instance, in the $\varphi$-coherent semantics, $ \varphi_i$ can be taken to be a two or three-valued approximation of the activation function of unit $i$.  


\section{A preferential interpretation of gradual argumentation semantics} \label{sec:Conditional_reasoning}

The strong relations between the notions of coherent, faithfull and $\varphi$-coherent labellings of a gradual argumentation graph and the 
corresponding semantics of weighted conditional knowledge bases have suggested 
   an approach for defeasible reasoning over a weighted argumentation graph \cite{WorkshopAI3_short,NMR2022}, 
   which builds on the semantics of the argumentation graph.
   


Given an argumentation graph $G$ and a gradual argumentation semantics $S$,
we define a preferential (many-valued) interpretation of the argumentation graph $G$, with respect to the gradual semantics $S$.
In the following, we first consider a propositional language to represent {\em boolean combination of arguments} and a many-valued semantics for it on the domain of argument valuation. 
Then, we extend the language with a {\em typicality operator}, to introduce defeasible implications over boolean combinations of arguments and define a {\em (multi-)preferential} interpretation associated with the argumentation graph $G$ and the argumentation semantics $S$.
A similar extension with typicality of a propositional language has, for instance, been considered for the two-valued case in the Propositional Typicality Logic 
\cite{BoothCasiniAIJ19}.

Let us consider an argumentation graph $G=\la \mathcal{A}, \mathcal{R}, \pi \ra$, which may be weighted or not, and some gradual argumentation semantics $S$ for $G$. 
In case the graph is not weighted, we let $\pi(B,A)=-1$ mean that argument $B$ attacks argument $A$, and $\pi(B,A)=+1$ mean that argument $B$ supports argument $A$.
We refer to \cite{BaroniRagoToni2018} for a classification and the properties of gradual semantics when the argumentation graph is not weighted.

We introduce a propositional language $\mathcal{L}$, whose set of propositional variables $Prop$ is the set of arguments $\mathcal{A}$.
Assume our language $\mathcal{L}$ contains the connectives $\wedge$, $\vee$, $\neg$ and $\rightarrow$, and that propositional formulas are defined inductively from the set of propositional variables, as usual. Propositional formulas, when restricted to the propositional variables in $\mathcal{A}$, correspond to {\em boolean combination of arguments}, which have been considered, for instance, by Hunter et al. \cite{HunterPolbergThimm_AIJ2020}. 
We will use $\alpha, \beta, \gamma$ to denote  boolean combination of arguments.

Let $\mathcal{S}$ be a {\em truth degree set}, equipped with a preorder relation $\leq$.
This choice for the domain of argument valuations has been proposed by Baroni et al. \cite{BaroniRagoToni2018} as it is general enough to include the domain of argument valuations in most gradual argumentation semantics.
%
Let the truth set $\mathcal{S}$ be the domain of argument valuation of a gradual semantics $S$ for the argumentation graph $G$.
%
%
%
We let $\otimes$, $\oplus$, $\rhd$ and $\ominus$ be 
the {\em truth degree functions} in $\mathcal{S}$ for the connectives $\wedge$, $\vee$, $\neg$ and $\rightarrow$ (respectively).
For instance, when  $\mathcal{S}$ is $[0,1]$ or the finite set  $\mathcal{C}_n$,  $\otimes$, $\oplus$, $\rhd$ and $\ominus$ can be chosed as a t-norm, s-norm, implication function, and negation function 
in some well known system of many-valued logic \cite{Gottwald2001}.

A labelling $\sigma: \mathcal{A} \rightarrow \mathcal{S}$ assigns to each argument $A_i \in \mathcal{A}$ a truth degree in $\mathcal{S}$, that is, 
$\sigma$ is a  {\em many-valued valuation}.  
%
A valuation $\sigma$ can be inductively extended to all propositional formulas of $L$ as follows: 
\begin{quote}
  $\sigma(\alpha \wedge \beta)= \sigma(\alpha) \otimes \sigma(\beta)$
  \ \ \ \ \ \ \ \ \ \ \ \ \ \ \ \ \ \ \ \ 
  $\sigma(\alpha \vee \beta)= \sigma(\alpha) \oplus \sigma(\beta)$
  
  $\sigma(\alpha \rightarrow \beta)= \sigma(\alpha) \rhd \sigma(\beta)$
  \ \ \ \ \ \ \ \ \ \ \ \ \ \ \ \ \ \ 
  $\sigma(\neg \alpha)= \ominus \sigma( \alpha)$
 \end{quote} 
Based on the choice of the combination functions, a labelling $\sigma$ uniquely assigns a truth degree to any  boolean combination of arguments. In the following we will assume that the false argument $\bot$ and the true argument $\top$  are formulas of $L$ and  that $\sigma(\bot)=0$ and $\sigma(\top)=1$, for all labellings $\sigma$.
 
Let $\Sigma$ be a set of labellings of a weighted argumentation graph $G=\la \mathcal{A}, \mathcal{R}, \pi \ra$, e.g., the labellings of $G$ under a given gradual semantics $S$ with domain of argument valuation $\mathcal{S}$.
As an example, $\Sigma$ may be the set of the finitely-valued $\varphi$-coherent labellings of the graph $G$ with values in $\mathcal{C}_n$.


Given a set of labellings $\Sigma$, for each argument $A_i \in \mathcal{A}$ , we can define a {\em preference relation $<_{A_i}$} on $\Sigma$, as follows:
$$\sigma <_{A_i} \sigma'   \mbox{ iff }   \sigma'(A_i) < \sigma(A_i), $$
where $<$ is the strict partial order on $\mathcal{S}$ defined from the preorder $\leq$ as usual
($a<b \mbox{ iff }  a \leq b \mbox{ and  } b \not\leq a$).
The labelling {\em $\sigma$ is preferred to  labelling $\sigma'$} with respect to argument $A_i$ (or $\sigma$ is {\em more plausible} than $\sigma'$ for argument $A_i$), when the degree of truth of $A_i$ in $\sigma$ is greater than  the degree of truth of $A_i$ in $\sigma'$.
The preference relation $<_{A_i}$ is a strict partial order relation on $\Sigma$.
When $\leq$ is a {\em total preorder} on $\mathcal{S}$,
$<_{A_i}$ is also {\em modular}, that is, for all $\sigma',\sigma'',\sigma''' \in \Sigma$, $\sigma' <_{A_i} \sigma''$ implies ($\sigma' <_{A_i} \sigma'''$ or $\sigma''' <_{A_i} \sigma''$).
For instance, when $\mathcal{S}$ is equal to $[0,1]$ or to $\mathcal{C}_n$,  $<_{A_i}$ is a strict modular partial order relation.

When the set $\Sigma$ is infinite,   $<_{A_i}$ is not guaranteed to be {\em well-founded}, as there may be infinitely-descending chains of labellings 
$\sigma_{2} <_{A_{i}} \sigma_{1}$, 
$\sigma_{3} <_{A_{i}} \sigma_{2}$, $\ldots$.
In the following, we will restrict our consideration to sets of labellings $\Sigma$ such that $<_{A_i}$ is well-founded for all arguments $A_i$ in $\Sigma$. We will call $\Sigma$ a {\em  well-founded} set of labellings.

The definition of preference over arguments which is induced by a set of labellings $\Sigma$ also extends to  boolean combination of arguments $\alpha$ in the obvious way,
 given a choice of combination functions. A set of labellings  $\Sigma$ {\em induces} a preference relation $<_{\alpha}$ on $\Sigma$, for each  boolean combination of arguments $\alpha$, as follows: $\sigma <_{\alpha} \sigma'   \mbox{ iff }    \sigma'(\alpha) < \sigma(\alpha) $.
 Note that if $<_{A_i}$ is well-founded for all arguments $A_i$ in $\sigma'$, it is well-founded for all boolean combination of arguments $\alpha$.

We now define a notion of preferential interpretation over many-valued valuations, which relates to preferential interpretations in KLM preferential logics and, when preferences $<_{A_i}$ are modular, to ranked interpretations \cite{whatdoes}.
As seen above, here preferences are defined over many-valued labellings, rather than over two-valued propositional valuations. 
A further difference with KLM approach is that the semantics exploits multiple preferences (one for each argument) rather than a single one, being then a {\em multi-preferential} semantics. Some multi-preferential semantics for KLM conditional logics have been considered in \cite{Delgrande2020,AIJ21},
and, for ranked and weighted knowledge bases in  fuzzy description logics in \cite{JELIA2021}. 
\begin{definition}
Given an argumentation graph $G=\la \mathcal{A},$  $\mathcal{R}, \pi \ra$,
and $\Sigma$ a set of labellings  of $G$, in some gradual semantics of $G$ with the domain of argument valuation $\mathcal{S}$,
a {\em preferential interpretation of  $G$ wrt. $\Sigma$}, is a pair $I=(\mathcal{S},\Sigma)$,
where $\Sigma$ is a set of labellings $\sigma: \mathcal{A} \rightarrow \mathcal{S}$.
\end{definition}
Note that preference relations $<_\alpha$ in $I$ are left implicit, as they are induced by the set of labellings in $\Sigma$.

A unary typicality operator $\tip$ can be added to the language $L$. We will call the extended language $L^\tip$, and the associated many-valued logic with typicality $\mathcal{L}^\tip$.
Intuitively, as in the Propositional Typicality Logic, ``a sentence of the form $\tip(\alpha)$ is understood to refer to the {\em typical situations in which $\alpha$ holds}" \cite{BoothCasiniAIJ19}.

The typicality operator allows the formulation of conditional implications (or {\em defeasible implications}) of the form $\tip(\alpha) \rightarrow \beta$, whose meaning is that "normally, if $\alpha$ then $\beta$", where $\alpha$ and $\beta$ are boolean combination of arguments. In the two-valued case such implications correspond to conditional implications $\alpha \ent \beta$ of KLM preferential logics \cite{whatdoes}.
As usual \cite{BoothCasiniAIJ19}, we do not allow nesting of the typicality operator.
When $\alpha$ and $\beta$ do not contain occurrences of the typicality operator, an implication $\alpha \rightarrow \beta$ is called {\em strict}. 
In the following, we do not restrict our consideration to strict or defeasible implications, but we allow in the language $L^\tip$ general implications of the form $\alpha \rightarrow \beta$, where $\alpha$ and $\beta$ are boolean combination of arguments which may contain (unnested) occurrences of the typicality operator. 

The interpretation of a typicality formula  $\tip(\alpha)$ is defined with respect to a preferential interpretation $I=(\mathcal{S},\Sigma)$.
We define the truth degree of  a typicality formula $\tip(\alpha)$ in a valuation $\sigma \in \Sigma$ of the preferential interpretation $I=(\mathcal{S},\Sigma)$ taking into account all labellings in $\Sigma$.
In the next definition we are assuming  the set of labellings $\Sigma$ to be well-founded.


\begin{definition}\label{defi:typicality}
Given a preferential interpretation $I=(\mathcal{S},\Sigma)$, and a labelling $\sigma \in \Sigma$, the valuation of a propositional formula $\tip(\alpha)$ in $\sigma$ is defined as follows:
\begin{align}\label{eq:interpr_typicality}
	\sigma(\tip(\alpha))  & = \left\{\begin{array}{ll}
						 \sigma(\alpha) & \mbox{ \ \ \ \  if } \sigma \in min_{<_\alpha} (\Sigma) \\
						0 &  \mbox{ \ \ \ \  otherwise } 
					\end{array}\right.
\end{align}  
where $min_{<_\alpha}(\Sigma)= \{\sigma: \sigma \in \Sigma$ and $\nexists \sigma' \in \Sigma$ s.t. $\sigma' <_\alpha \sigma \}$.
\end{definition}
The truth value of $\tip(\alpha)$ in $\sigma \in \Sigma$ depends on the whole set of labellings $\Sigma$ in the preferential interpretation $I$. 
When $(\tip(A))^I(\sigma)>0$, $\sigma$ is 
a labelling assigning a maximal degree of acceptability to  argument $A$ in $I$.
i.e., it is a labelling maximizing the acceptability of argument $A$, among all the labellings in $I$. 
Similarly, in Propositional Typicality Logic \cite{BoothCasiniAIJ19}, the evaluation of a typicality formula in a preferential interpretation depends on the set of all propositional valuations in the interpretation.

%

Observe that, with the definition above, we are departing from previous semantics of typicality in the fuzzy and many-valued description logics  \cite{JELIA2021,ICLP22}, as we are {\em not} defining a two-valued interpretation for $\tip(A)$. The typicality operator is not simply intended to single out those labellings $\sigma \in \Sigma$ which maximize the acceptability degree of argument $A$, 
but it assigns a truth degree to $\tip(A)$ in $\sigma$ which may be different from $0$ or $1$.



Given a preferential interpretation $I=(\mathcal{S},\Sigma)$, where the labellings in $\Sigma$ have domain of argument valuation $\mathcal{S}$ we can now define the satisfiability in $I$ of a {\em graded implication}, having the form $\alpha  \rightarrow \beta \geq l$ or $\alpha  \rightarrow \beta \leq u$, for $l$ and $u$ values in $\mathcal{S}$.
We first define the truth degree of an implication $\alpha \rightarrow \beta$ wrt. a preferential interpretation $I$.
\begin{definition}
Given a preferential interpretation  $I=(\mathcal{S},\Sigma)$ associated with an argumentation graph $G$,
the {\em truth degree of an implication $\alpha \rightarrow \beta$ wrt. $I$} is defined as follows:
\begin{quote}
$(\alpha \rightarrow \beta)^I= inf_{\sigma \in \Sigma} (\sigma(\alpha) \rhd \sigma(\beta) )$
\end{quote}
\end{definition}
As a special case, for defeasible implications, we have that:
\begin{quote}
$(\tip(\alpha) \rightarrow \beta)^I= inf_{\sigma \in \Sigma} (\sigma(\tip(\alpha)) \rhd \sigma(\beta) )$
\end{quote}
meaning that in the most typical situations (labellings) in which $\alpha$ holds, $\beta$ also holds.
Note that the interpretation of an implication (and of a defeasible implication) is defined {\em globally} wrt. a preferential interpretation $I$, and it is based on the whole set of labellings $\Sigma$ in $I=(\mathcal{S},\Sigma)$. This is similar to the interpretation of conditionals $A \ent B$ in a preferential model in KLM preferential semantics \cite{KrausLehmannMagidor:90,whatdoes}, as the semantics of conditional formulas depends (globally) on the set of propositional valuations in the preferential model as well as on the preference relation.

We can now define the satisfiability of a {\em graded implication} in a preferential interpretation $I=(\mathcal{S},\Sigma)$. 
\begin{definition}
Given a preferential interpretation $I=(\mathcal{S},\Sigma)$ of an argumentation graph $G$,
 {\em $I$ satisfies a graded implication $\alpha  \rightarrow \beta \geq l$} (written $I \models \alpha  \rightarrow \beta \geq l$) iff $(\alpha  \rightarrow \beta)^I \geq l$;
$I$ satisfies a graded implication $\alpha  \rightarrow \beta \leq u$ (written $I \models \alpha  \rightarrow \beta \leq u$) iff $(\alpha  \rightarrow \beta)^I \leq u$.
\end{definition}

Notice that the valuation of a graded implication (e.g., $\alpha  \rightarrow \beta \geq l$) in a preferential interpretation $I$ is two-valued (that is, either the graded implication is satisfied in $I$ (i.e., $I \models \alpha  \rightarrow \beta \geq l$) or it is not (i.e., 
$I \not \models \alpha  \rightarrow \beta \geq l$). Hence, it is natural to consider {\em boolean combinations of graded 
implications}, such as the following ones: 

$(\tip(A_1) \rightarrow A_2 \wedge A_3 \leq 0.3) \wedge (\tip(A_3 \vee A_1) \rightarrow A_4) \geq 0.6)$, and

$(\tip(A_1) \rightarrow A_2 \wedge A_3 \leq 0.3) \rightarrow (\tip(A_1) \rightarrow A_4) \geq 0.6)$,

\noindent
and define their satisfiability in an interpretation $I$ in the obvious way, based on the semantics of classical propositional logic.
For two graded implications, $\psi$ and $\eta$, we have:
\begin{quote}
 $I \models \psi \wedge \eta$ iff $I \models \psi $ and $I \models \eta$;
 
$I \models \psi \vee \eta$ iff $I \models \psi $ or $I \models \eta$;


$I \models \neg \psi$ iff $I \not \models \psi$
\end{quote}
and similarly for all other  classical logic connectives\footnote{Here, we have used the same names for the boolean connectives $\wedge, \vee, \neg \rightarrow$ over graded implications and for the many-values connectives $\wedge, \vee, \neg \rightarrow$ over boolean combination of arguments. 
However, there is no ambiguity on their interpretation, which only depends on their occurrence.}. 



Given a specific semantics $S$ of a weighted argumentation graph $G$, with domain of argument valuation $\mathcal{S}$, if  the set of labellings $\Sigma$  of $G$ in $S$ is well-founded, we let  $I^S=(\mathcal{S}, \Sigma)$ be the preferential interpretation of $G$ with respect to the gradual semantics $S$.

Some gradual semantics assume a {\em base score} (or basic weight) $\sigma_0$, which is a function 
$\sigma_0: \mathcal{A} \rightarrow \mathcal{S}$ assigning to each argument $A \in \mathcal{A}$ its basic strength.
A gradual semantics can then be defined with respect to an argumentation graph and a base score $\sigma_0$ (see \cite{Amgoud2017,BaroniRagoToni2018,Amgoud2019} for the description of general frameworks and references to the many proposals). For such argumentation semantics $S$, a set of labellings $\Sigma^S_{\sigma_0}$ can be associated to a given choice $\sigma_0$ of the base score, and 
a preferential interpretation 
$I^S=(\mathcal{S}, \Sigma^S)$ can be defined for a semantics $S$ by considering the labellings for {\em all} the possible choices of the base score  $\sigma_0$, or for {\em some} of them (the choices satisfying some conditions), or for a single one.

The preferential interpretation $I^S$ 
can be used to validate (under the semantics $S$) properties of interest of an  argumentation graph $G$, expressed by graded implications (including strict or defeasible implications or their boolean combination) based on the semantics $S$. 
When the preferential interpretation $I^S$ 
is finite (i.e., contains a finite set of labellings), the satisfiability 
of graded implications (or their boolean combinations) can be verified by {\em model checking} over the preferential interpretation $I^S$. 
%
In case there are infinitely many labellings of the graph in the semantics $S$, which may give rise to non well-founded preference relations associated to arguments, approximations of the semantics $S$ over a finite domain, 
can be considered for proving properties of the argumentation graph.

In the next section, 
as a proof of concept, we consider the model checking problem for the argumentation semantics based on $\varphi$-coherent finitely-valued labellings introduced in  Section \ref{sec:varphi_labellings} 
in the case $\mathcal{S}= \mathcal{C}_n$,  for an integer $n\geq 1$.

\section{An ASP approach for conditional reasoning 
in the finitely-valued case} \label{sec:ASP} 

In this section, as a proof of concept, we describe an ASP approach for conditional reasoning over an argumentation graphs based on a finitely-valued semantics.
We consider the $\varphi$-coherent finitely-valued semantics of a weighted argumentation graph $G$ introduced in Section \ref{sec:varphi_labellings}, with domain of argument valuation $\mathcal{C}_n$, for some integer $n\geq 1$. We will denote the set of labellings in this argumentation semantics
$\Sigma^G_{\varphi,n}$.

The idea is that of representing in ASP  a many-valued labelling as an Answer Set which records the assignment of a value in $\mathcal{C}_n$ to each argument $A_i$. The labelling is encoded by a set of atoms of the form $\mathit{val(a,v)}$, meaning that $\frac{v}{n} \in C_n$ is the acceptability degree of argument $a$ in the labelling.
The rule:  

$\;$ \ \  $ \mathit{
1\{val(A, V) : val(V)\}1 \ \leftarrow arg(A).  
	} $

\noindent	
where facts $ \mathit{val(v)}$ and $ \mathit{arg(a)}$ hold for all $v$ s.t.\ $\frac{v}{n} \in C_n$ and  $ a \in \mathcal{A}$,
generates candidate answer sets, corresponding to the different labellings, with all the possible choices of the values  $\frac{v}{n} \in \mathcal{ C}_n$  in each argument $A$.  

The valuation of boolean combinations $B$ of arguments is encoded as a predicate $\mathit{eval(B, V)}$. A rule is introduced for each connective  to encode its semantics, based, e.g.,  on G\"odel logic with standard involutive negation:
$a \otimes b= min\{a,b\}$,  $a \oplus b= max\{a,b\}$,  $a \rhd b= 1$ {\em if} $a \leq b$ {\em and} $b$ {\em otherwise}; and $ \ominus a = 1-a$
(but other choices of combination functions can as well be considered). 
Predicate $\mathit{arg\_comb}$ is used to limit the instantiation of rules (in the {\em grounding} before ASP {\em solving})
to boolean combinations of interest, which are the ones used in queries and their subformulae (due to rules such as $\mathit{arg\_comb(A)  \leftarrow  arg\_comb(and(A, B)) }$.
Then we have: 


$\;$ \ \ $ \mathit{
eval(A, V) \leftarrow arg(A), val(A,V)}$.


$\;$ \ \   $ \mathit{
eval(and(A, B), V) \leftarrow arg\_comb(and(A, B)), eval(A,V1), }$  

$\;$ \ \ \ \ \ \ \ \ \ \ \ \    \ \ \ \ \ \ \ \ \ \ \ \   \ \ \ \ \ \ \ \ \ \ \ \  $ \mathit{eval(B,V1), min(V1, V2,V).
	} $
	
$\;$ \ \   $ \mathit{
eval(or(A, B), V) \leftarrow arg\_comb(or(A, B)), eval(A,V1),  }$  

$\;$ \ \ \ \ \ \ \ \ \ \ \ \    \ \ \ \ \ \ \ \ \ \ \ \   \ \ \ \ \ \ \ \ \ \ \ \  $ \mathit{eval(B,V1),  max(V1, V2,V).
	} $

$\;$ \ \   $ \mathit{
eval(impl(A, B), n) \leftarrow arg\_comb(impl(A, B)), eval(A,V1),   }$ 

$\;$ \ \ \ \ \ \ \ \ \ \ \ \    \ \ \ \ \ \ \ \ \ \ \ \   \ \ \ \ \ \ \ \ \ \ \ \  $ \mathit{ eval(B,V2), V1 \leq V2.
	} $

$\;$ \ \   $ \mathit{
eval(impl(A, B), V2) \leftarrow arg\_comb(impl(A, B)), eval(A,V1),   }$ 

$\;$ \ \ \ \ \ \ \ \ \ \ \ \    \ \ \ \ \ \ \ \ \ \ \ \   \ \ \ \ \ \ \ \ \ \ \ \  $ \mathit{ eval(B,V2), V1 > V2.
	} $





$\;$ \ \ $ \mathit{
eval(neg( A), V) \leftarrow arg\_comb(neg (A)), eval(A,V1),  
	} $
	
$\;$ \ \ \ \ \ \ \ \ \ \ \ \    \ \ \ \ \ \ \ \ \ \ \ \   \ \ \ \ \ \ \ \ \ \ \ \  $  \mathit{V= n-V1.}$

\noindent	
where $\mathit{arg}$ holds for arguments, and $\mathit{min}$ 
and  $\mathit{max}$ 
are suitably defined.

To encode the $\varphi$-coherent semantics we need to encode condition (\ref{cond_phi_coherence}). 
Even if the computation of the weighted sum and of $\varphi$
is expressible in terms of ASP rules, the evaluation of such rules would eventually materialize all combinations of truth degrees for all attacker arguments, which is practically feasible only if the maximum in-degree is bounded by a small number.
Moreover, such ASP rules would require to represent weights on the graph in terms of integers, hence introducing an approximation.
%
We opted for an alternative approach powered by the Propagator interface of the {\em clingo} API \cite{Gebser16}.
In a nutshell, we defined a custom propagator that enforces the condition 

 $\mathit{\bot \leftarrow  arg(A), eval(A,V), V != \varphi(V / n)}.$

\noindent
for all arguments $A \in \mathcal{A}$ s.t.\ $R^-(A) \neq \emptyset$.
The propagator takes as input the argument $A$, and is initialized after the grounding phase, when it can identify all attackers/supporters of $A$, the associated weights, and set watches for the boolean variables that \emph{clingo} associates to the instances of predicate $\mathit{eval}$.
After that, whenever a watched boolean variable is assigned true or unrolled to undefined, the propagator is notified and keeps track of the change.
If all attackers/supporters of $A$ have been assigned a truth degree, then the propagator can infer the truth degree of $A$, and provide an explanation to \emph{clingo} for such an inference in terms of a clause.
Similarly, if $A$ has already a truth degree and all attackers/supporters of $A$ have been assigned a truth degree which is incompatible with that of $A$, then the propagator can report a conflict, and provide an explanation to \emph{clingo} for such a conflict in terms of a clause.

The preferential interpretation $I^G_{\varphi,n} =(\mathcal{C}_n, \Sigma^G_{\varphi,n})$  which is built over the (finite) set $\Sigma^G_{\varphi,n}$ of all the labellings  of the argumentation graph $G$ in the $\varphi$-coherent finitely-valued semantics,
is represented by all resulting answer sets.


We consider the evaluation of defeasible and strict graded implications.
For the verification that a defeasible graded implication $\tip(\alpha) \rightarrow \beta \geq l$ is satisfied  over the preferential interpretation $I^G_{\varphi,n}$ of the graph $G$,
one has to check that in all the labellings $\sigma$ that maximize the value of $\sigma(\alpha)$ wrt. all labellings in  $\Sigma^G_{\varphi,n}$, it holds that
$\sigma(\tip(\alpha)) \rhd \sigma(\beta) \geq l$.
That is, one has to verify that in all the answer sets $M$ that maximize the value of $V$ such that $\mathit{eval(\alpha,V) \in M}$, it holds that, if
$\mathit{eval(\alpha,v1)}$, $\mathit{eval(\beta,v2) \in M}$, then $v1 \leq v2$ or $v2 \geq l$ also hold in $M$.
A counterexample can be searched, i.e.\ an answer set that maximizes the evaluation $v1$ of $\alpha$, and is such that, if $v2$ is the evaluation of $\beta$, $v1 > v2$ and $v2 < l$.

If the defeasible graded implication $\tip(\alpha) \rightarrow \beta \geq l$
is represented as 
$\mathit{query(typ(\alpha'),}$ $\mathit{\beta',l'), }$ (where $\mathit{\alpha',\beta'}$ represent 
$\mathit{\alpha,\beta}$ and
$l'= l*n$), the weak constraints:

\smallskip
$\mathit{
:\sim query(typ(Alpha),\_,\_), eval(Alpha,V). [-1@V+2, Alpha,V]
} $

\noindent
prefer answer sets where the evaluation of
$\alpha$ is maximal, given that the presence of $\mathit{eval(Alpha,V)}$ has priority $V+2$, then increasing with $V$.

The preference for a counterexample is obtained with the additional weak constraint:

\smallskip
$\mathit{
:\sim query(typ(Alpha),Beta,L), eval(impl(Alpha,Beta),V), V<L. }$

$\;$ \ \ \ \ \ \ \ \ \ \ 
$ \mathit{ [-1@1, Alpha,Beta,L,V1,V2]
} $
\smallskip

\noindent
which is given priority 1, smaller than the previous ones. In fact, in the solutions selected by the higher priority constraints, the evaluation of $\tip(\alpha) \rightarrow \beta$ coincides with the one for 
$\alpha \rightarrow \beta$.

The verification of a strict graded implication of the form $\alpha \rightarrow \beta \geq l$ is simpler, as it does not involve maximizing the evaluation of a formula.

\begin{figure}[t]
	\centering
	\includegraphics[width=0.4\textwidth]{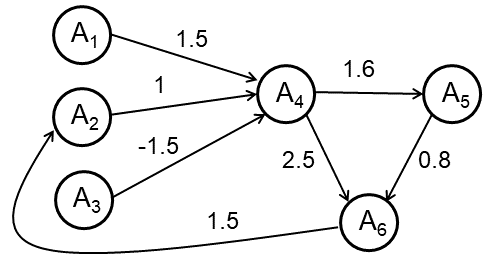}
	\caption{ Example weighted argumentation graph \label{arggraph} }
	
\end{figure}

As an example, for the weighted argumentation graph in Figure \ref{arggraph}, there are 36 labellings in case $n=5$ and $\varphi$ is the approximation of the sigmoid function to the closest value in $\mathcal{C}_n$.  
The following graded implications hold:
$\tip(A_1 \wedge A_2 \wedge \neg A_3) \rightarrow A_6 \geq 1$ (with 4 preferred labellings), $\tip(A_1 \wedge A_2) \rightarrow A_6 \geq 4/5$,
$\tip(A_1 \wedge A_2) \rightarrow A_6 \geq 4/5$ (12 preferred labellings), 
$\tip(A_6) \rightarrow A_1 \wedge A_2 \geq 4/5$ (1 preferred labelling).
On the other hand, $A_6 \rightarrow A_1 \wedge A_2 \geq 1/5$ does not hold.


\section{Towards a probabilistic semantics of gradual argumentation} \label{sec:probabilistic_semantics}


When the domain of argument valuation is in the interval $[0,1]$, the definition of a preferential interpretation $I^S$ associated to a gradual semantics $S$, which has been developed in Section \ref{sec:Conditional_reasoning}, also suggests a probabilistic argumentation semantics, inspired to Zadeh's {\em probability of fuzzy events} \cite{Zadeh1968}. 
The approach has been considered 
for providing a probabilistic interpretation of Self-Organising Maps after training \cite{JLC2022}, by  exploiting a recent characterization of the continuous t-norms compatible with Zadeh's probability of fuzzy events ($P_Z$-compatible t-norms) by Montes et al.\ \cite{Montes2013}.

Montes et al.\ \cite{Montes2013} have  
studied the problem of determining which t-norms make Zadeh's probability of fuzzy events fulfill Kolmogorov's axioms 
(called $P_Z$-compatible t-norms). 
For a given t-norm $T$, they consider a generalization of the notion of algebra to the fuzzy framework by means of T-clans\footnote{
A T-clan over $\Omega$ is 
a subset $\mathcal{ B}$ of $\mathcal{ F}(\Omega)$ satisfying the conditions: 
(i) $\emptyset \in \mathcal{ B}$; (ii) if $A \in \mathcal{ B}$, then its complement $A^c \in \mathcal{ B}$;
(iii) if $A , B \in \mathcal{B}$, then $A \cap_{T} B \in \mathcal{ B}$, where, for a t-norm $T$, $(A \cap_{T} B)(w)= T(A(w),  B(w))$ for all $w \in \Omega$ \cite{Montes2013}.}.
Given the universe $\Omega$, the set $\mathcal{ F}(\Omega)$ of all fuzzy subsets on $\Omega$, and a probability measure $P$ over $\Omega$,
they define the probability of a measurable fuzzy event $A$ in a {\em T-clan} over $\Omega$. 


In this section we explore this approach in the context of gradual argumentation, to see that it leads to a generalization of the probabilistic semantics presented in \cite{ThimmECAI2012}, and we discuss some advantages and drawbacks of the approach.


Let $S$ be some gradual argumentation semantics of an argumentation graph $G=\la \mathcal{A}, \mathcal{R}, \pi \ra$, which may be weighted or not (as considered in section \ref{sec:Conditional_reasoning}). Let $\Sigma^S$ be a set of labellings of $G$ in a gradual semantics $S$ (which may depend or not on a base score $\sigma_0$ of the arguments). Let us assume that $\mathcal{S}=[0,1]$ is the domain of argument valuation in $S$ and that $I^S=(\mathcal{S}, \Sigma^S)$ is the preferential interpretation associated to $S$, as defined in Section \ref{sec:Conditional_reasoning}.

Let $\Sigma$ be the set of labellings $\Sigma^S$ of $G$ in the gradual argumentation semantics $S$, or a subset of it.
For instance, we may take $\Sigma$ to be the set of $\varphi$-coherent labellings, either in the infinitely-valued or in the finitely-valued case. 

The probabilistic semantics we propose, is inspired to  Zadeh's {\em probability of fuzzy events} \cite{Zadeh1968}, as one can regard an argument $A \in \mathcal{A}$ as a fuzzy event, with membership function $\mu_A: \Sigma \rightarrow [0,1]$, where $\mu_A(\sigma)=\sigma(A)$.
Similarly, any boolean combination of arguments $\alpha$ can as well be regarded as a fuzzy event, 
with membership function $\mu_\alpha(\sigma)=\sigma(\alpha)$, 
where the extension of labellings to boolean combinations of arguments 
has been defined in Section  \ref{sec:Conditional_reasoning}. 

Let us restrict to a $P_Z$-compatible t-norm $\otimes$, with associated  t-conorm $\oplus$ and the negation function $\ominus x= 1-x$.
For instance, one can take the minimum t-norm, Product t-norm, or Lukasiewicz t-norm.
Let  $\mathcal{ F}(\Sigma)$ denote the set of all fuzzy subsets on $\Sigma$. 
Given $I^{S}=\langle \mathcal{S}, \Sigma \rangle$, 
each boolean combination of arguments $\alpha$ is interpreted as a fuzzy event on $\Sigma$, with membership function 
$\mu_\alpha: \Sigma \ri [0,1]$, defined as above (that we assume to be measurable).
The collection of the fuzzy events, associated to the boolean combination of arguments $\alpha$, forms a $T$-clan over $\Sigma$. 

Given $I^{S}=\langle \mathcal{S}, \Sigma \rangle$,
we assume a discrete probability distribution $p: \Sigma \rightarrow [0,1]$ over the set $\Sigma$,
and define the {\em probability of a boolean combination of arguments 
$\alpha$} as follows:
\begin{equation} \label{def: Probability_bool_arg}
P(\alpha)=\sum_{\sigma \in \Sigma} \sigma(\alpha) \; p(\sigma)
\end{equation}
%
For a single argument $A \in \mathcal{A}$, when labellings are two-valued (that is, $\sigma(A)$ is $0$ or $1$), the definition above becomes the following:
$P(A)=\sum_{\sigma \in \Sigma \wedge \sigma(A)=1} p(\sigma)$,
which relates to the probability of an argument in the probabilistic semantics 
in \cite{ThimmECAI2012}. 
In \cite{ThimmECAI2012} the probability of an argument $A$ in $Arg$ is ``the degree of belief that $A$ is in an extension", defined as the sum of the probabilities of all possible extensions $e$ that contain argument $A$, i.e.,
 $P(A)=\sum_{A \in e \subseteq Arg} \; p(e),$
where an extension $e \in 2^{Arg}$ is a set of arguments in $Arg$, and $p(e)$ is the probability of extension $e$.
Here, on the other hand, we are considering many-valued  labellings assigning an acceptability degree $\sigma(A)$ to arguments, so it may be not the case that an argument either belongs to an extension (a labelling) or it does not.

Following Smets \cite{Smets82}, 
we let the {\em conditional probability of $\alpha$ given $\beta$}, where $\alpha$ and $\beta$ are boolean combinations of arguments,
to be defined as 
$$P(\alpha | B)=P(\alpha \wedge \beta) / P(\beta)$$ (provided $P(\beta)>0$). 
As observed by Dubois and Prade \cite{DuboisP1993_fuzzy_probab}, this generalizes both conditional probability and the fuzzy inclusion index advocated by Kosko \cite{Kosko92}. 

It can be shown that $\sigma(A)$ can be interpreted as the conditional probability of argument $A$, given labelling $\sigma$, and it can be regarded as a subjective probability (a degree of belief we put into $A$ when we are in the state represented by labelling $\sigma$).
Let us consider the language $L^\tip$ extended by introducing a new proposition $\{\sigma\}$, for each $\sigma \in \Sigma$.
Let us also extend the valuations $\sigma$ to such propositions by letting:
$\sigma(\{\sigma\})=1$ and $\sigma'(\{\sigma\})=0$, for any $\sigma' \in \Sigma$ such that $\sigma' \neq \sigma$.
It can be proven that the collection of the fuzzy events, associated to the boolean combination of arguments $\alpha$, in the extended language including the typicality operator and the new propositions $\{\sigma\}$, forms a $T$-clan over $\Sigma$. 
Furthermore, it can be shown (similarly to \cite{JLC2022}) that 
$$P(A | \{\sigma\})= \sigma(A).$$ 
The result holds when the t-norm is chosen, e.g., as in G\"odel, \L ukasiewicz or Product logic.
Then, $\sigma(A)$ can be interpreted as the conditional probability that argument $A$ holds, given labelling $\sigma$.

Under the assumption that the probability distribution $p$ is {\em uniform} over the set $\Sigma$ of labellings,  it can be proven that $P(\alpha | \beta)= M(\alpha \wedge \beta)/M(\beta)$,  where $M(\alpha)= \sum_{x \in \Sigma} \sigma(A)$ is the {\em size} of the fuzzy event $\alpha$ (provided $M(\beta)>0$).

Starting from an argumentation graph $G$ and a gradual semantics $S$ for it, if the set of labellings  $\Sigma$  of $G$ in $S$ (possibly, wrt. a choice of basic strength functions) is finite, one can then compute 
(for some probability distribution $p$ over $\Sigma$) the probability of an argument (or a boolean combination of arguments) with respect to the given argumentation semantics $S$, as well as conditional probabilities.
For a finite set of labellings $\Sigma^S= \{\sigma_1, \ldots, \sigma_m\}$ wrt. a given semantics $S$, assuming a uniform probability distribution, 
we have that
$P(\alpha)= (\sigma_1(\alpha) + \ldots + \sigma_m(\alpha)) /m = M(\alpha)/m $, for any boolean combination of arguments $\alpha$.

For the example in Figure \ref{arggraph}, for the $\varphi$-coherent semantics  with truth degree set $\mathcal{C}_5$, assuming a uniform probability distribution, $P(A_1)= 0.5$, $P(A_2)=0.777$, $P(A_3)=0.483$, $P(A_4)=0.644$, $P(A_5)=0.222$, $P(A_6)=0.755$.
Furthermore, $P(A_1 \wedge A_2| A_6)=0.618$ and 
$P(A_1 \wedge A_2 \wedge \neg A_3| A_6)=0.397$,
while $P(A_1 \wedge A_2| \tip(A_6))=0.8$ and $P(A_1 \wedge A_2 \wedge \neg A_3| \tip( A_6))=0.8$. 

%
While the notion of probability $P$ defined by equation (\ref{def: Probability_bool_arg}) satisfies Kolmogorov's axioms 
for any $P_Z$-compatible t-norm, with associated  t-conorm, and the negation function $\ominus x= 1-x$ \cite{Montes2013},
there are properties of classical probability, which may not hold (depending on the choice of t-norm).
This is a consequence of the fact that not all classical logic equivalences hold in a fuzzy logic.

For instance, the truth degree of 
$A \wedge \neg A$ in a labelling $\sigma$ may be different from $0$ depending on the t-norm 
(e.g., with G\"oedel and Product t-norms). Hence, it may be the case that 
$P(A \wedge \neg A)$ is different from $0$. 
Similarly, it may be the case that $P(A \vee \neg A)$ is different from $1$ (e.g., with G\"oedel t-norm) and that $P(A | A)$ is different from $1$ (e.g., with Product t-norm). 
While $P(A) + P(\neg A)=1$ holds (due to the choice of negation function),
$P(A|B)+P(\neg A|B)$ may be different from $1$.

In spite of the simplicity of this approach, on the negative side, some properties of classical probability are lost. Hence, we can consider the proposal in this section as a first step towards a probabilistic semantics for gradual argumentation.



 

\section{Conclusions}

In this paper 
we have developed a general approach to define a many-valued preferential interpretation of an argumentation graph, based on a gradual argumentation semantics. 
The approach allows for graded strict and defeasible implications involving arguments and boolean combination of arguments (with typicality) to be evaluated.
A preferential interpretation $I^S$ of the argumentation graph can be defined, based on some  chosen gradual argumentation semantics $S$. 
When the preferential interpretation is finite, the validation of (strict or defeasible) graded implications can be done by model-checking over the preferential interpretation. 
As a proof of concept, the paper has presented an Answer set Programming approach for conditional reasoning in the $\varphi$-coherent argumentation semantics in the finitely-valued case. 
The paper also proposes a probabilistic semantics of gradual argumentation, for a class of gradual semantics,
by adapting to argumentation the idea of Zadeh's probability of fuzzy events. 
In Section \ref{sec:probabilistic_semantics},  we have discussed this approach with its advantages and limitations.


Concerning the relationships between argumentation semantics and conditional reasoning, Weydert \cite{Weydert2013} has proposed one of the first approaches for combining abstract argumentation with a conditional semantics.
He has studied ``how to interpret abstract argumentation frameworks by instantiating the arguments and characterizing the attacks with suitable sets of conditionals describing constraints over ranking models". In doing this he exploits the JZ-evaluation semantics, which is based on system JZ \cite{Weydert03}.
Our approach aims to provide a preferential and conditional interpretation for a class of gradual argumentation semantics.

A correspondence between Abstract Dialectical Frameworks  \cite{Brewka2013} and Nonmonotonic Conditional Logics has been studied in
\cite{IsbernerFLAIRS2020},    with respect to the two-valued models, the stable, the preferred semantics and the grounded semantics of ADFs.

In \cite{SkibaThimm2022} 
Ordinal Conditional Functions (OCFs) are interpreted and formalized for Abstract Argumentation,
by developing a framework that allows to rank sets of arguments wrt. their plausibility. 
An attack from argument a to argument b is interpreted as the conditional relationship “if a is acceptable then b should not be acceptable”.
Based on this interpretation, an OCF inspired by System Z ranking function is defined.


In \cite{WorkshopAI3_short,NMR2022} 
an approach is presented which regards a weighted argumentation graph as a weighted conditional knowledge base in a fuzzy defeasible Description Logic. In this approach, a pair of arguments $(B,A) \in \mathcal{R}$ with weight $w_{AB}$ (representing an attack or a support), corresponds to a conditional implication $\tip(A) \sqsubseteq B$ with weight $w_{AB}$. Based on this correspondence, some semantics for weighted knowledge bases with typicality 
\cite{JELIA2021} 
have inspired some argumentation semantics \cite{WorkshopAI3_short}, and vice-versa. 

In \cite{ThimmECAI2012} a probabilistic semantics for abstract argumentation is proposed that assigns probabilities (or degrees of belief) to individual arguments.
As we have seen in Section \ref{sec:probabilistic_semantics}, the probability of an argument is defined as the sum of the probabilities of all possible extensions that contain the argument.
A notion of p-justifiable probability function of an Argumentation Framework (AF) is defined, which generalizes the notion of complete labelling to the probabilistic setting.
The paper also  investigates the relationships between classical argumentation semantics and probabilistic semantics.
It proves that the grounded labeling of an AF corresponds to the maximum entropy model of all p-justifiable probability
functions, and that the set of stable labellings of an AF (when non-empty) corresponds to the set of minimum entropy models of p-justifiable probability functions. 


In Section \ref{sec:probabilistic_semantics}, we have proposed a probabilistic semantics for argumentation, which  builds on a class of gradual argumentation semantics and is inspired to Zadeh's {\em probability of fuzzy events} \cite{Zadeh1968}.
We have seen that it can be regarded as a generalization of the probabilistic semantics  by Thimm \cite{ThimmECAI2012} and that
the approach allows the truth degree $\sigma(A)$ of an argument $A$ in a labelling $\sigma$ to be regarded as the conditional probability of $A$ given $\sigma$. On the other hand, 
as we have seen, some classical equivalences may not hold (depending on the choice of combination functions), and some properties of classical probability may be lost. This requests for further investigation. 
Alternative approaches, such as the one proposed recently by Flaminio et al. \cite{FlaminioGodo2020} for combining conditionals and probabilities,
might suggest for alternative ways of defining a probabilistic semantics for gradual argumentation.  




While in the paper we have followed an epistemic approach to probabilistic argumentation \cite{ThimmECAI2012,HunterPolbergThimm_AIJ2020}, 
in the constellation approach, the uncertainty resides in the topology of the Argumentation Framework, and a probability distribution over the sub-graphs of the argument graph is introduced  \cite{Bistarelli2022,LiOren2011}.

\end{document}